\documentclass[sigconf]{acmart}

\copyrightyear{2022}
\acmYear{2022}
\setcopyright{acmcopyright}\acmConference[KDD '22]{Proceedings of the 28th ACM SIGKDD Conference on Knowledge Discovery and Data Mining}{August 14--18, 2022}{Washington, DC, USA}
\acmBooktitle{Proceedings of the 28th ACM SIGKDD Conference on Knowledge Discovery and Data Mining (KDD '22), August 14--18, 2022, Washington, DC, USA}
\acmPrice{15.00}
\acmDOI{10.1145/3534678.3539443}
\acmISBN{978-1-4503-9385-0/22/08}

\usepackage[utf8]{inputenc}

\usepackage{latexsym}
\usepackage{microtype}
\usepackage{graphicx}
\usepackage{enumitem}

\usepackage{subfigure}
\usepackage{bbm}
\usepackage{array}
\usepackage{makecell}
\usepackage{subfigure}
\usepackage{xspace}
\newcommand{\mask}{\texttt{[MASK]}\xspace}
\newcommand{\model}{\textsf{ALIGNIE}\xspace}
\usepackage{amsmath}
\usepackage{amsfonts}
\usepackage{wrapfig}
\usepackage{multirow}
\usepackage{mathtools} 

\usepackage{verbatim}

\usepackage{anyfontsize}

\usepackage{color}
\usepackage{tikz}
\usepackage{adjustbox}

\newcommand{\ie}[0]{\emph{i.e., }}

\newcommand{\eg}[0]{\emph{e.g., }}

\newcommand{\etc}[0]{\emph{etc. }}

\newcommand{\Emat}[0]{{{\bf E}}}

\newcommand{\Umat}[0]{{{\bf U}}}

\newcommand{\Wmat}[0]{{{\bf W}}}

\newcommand{\bv}[0]{{\boldsymbol{b}}}

\newcommand{\hv}[0]{{\boldsymbol{h}}}

\newcommand{\mv}{{\boldsymbol{m}}}

\newcommand{\tv}[0]{{\boldsymbol{t}}}
\newcommand{\uv}{\boldsymbol{u}}

\newcommand{\xv}{\boldsymbol{x}}
\newcommand{\yv}{\boldsymbol{y}}

\newcommand{\thetav}{\boldsymbol{\theta}}

\newcommand{\R}{\mathbb{R}}

\newcommand{\Ycal}{\mathcal{Y}}
\newcommand{\Lcal}{\mathcal{L}}

\newcommand{\Dcal}{\mathcal{D}}

\newcommand{\Vcal}{\mathcal{V}}

\usepackage{tabularx,caption}
\usepackage[leftcaption]{sidecap}
\usepackage{booktabs}

\begin{document}
\fancyhead{}

\title{Few-Shot Fine-Grained Entity Typing with Automatic Label Interpretation and Instance Generation}
\author{Jiaxin Huang}
\affiliation{
\institution{jiaxinh3@illinois.edu }
\institution{University of Illinois at Urbana-Champaign}
\city{Urbana}\state{IL}\country{USA} 
}

\author{Yu Meng}
\affiliation{
\institution{yumeng5@illinois.edu }
\institution{University of Illinois at Urbana-Champaign}
\city{Urbana}\state{IL}\country{USA} 
}
\author{Jiawei Han}
\affiliation{
\institution{hanj@illinois.edu }
\institution{University of Illinois at Urbana-Champaign}
\city{Urbana}\state{IL}\country{USA} 
}

\date{February 2022}

\begin{abstract}
We study the problem of few-shot Fine-grained Entity Typing (FET), where only a few annotated entity mentions with contexts are given for each entity type.
Recently, prompt-based tuning has demonstrated superior performance to standard fine-tuning in few-shot scenarios by formulating the entity type classification task as a ``fill-in-the-blank'' problem. This allows effective utilization of the strong language modeling capability of Pre-trained Language Models (PLMs). 
Despite the success of current prompt-based tuning approaches, two major challenges remain: (1) the verbalizer in prompts is either manually designed or constructed from external knowledge bases, without considering the target corpus and label hierarchy information, and (2) current approaches mainly utilize the \textit{representation power} of PLMs, but have not explored their \textit{generation power} acquired through extensive general-domain pre-training.
In this work, we propose a novel framework for few-shot FET consisting of two modules: (1) an entity type label interpretation module automatically learns to relate type labels to the vocabulary by jointly leveraging few-shot instances and the label hierarchy, and (2) a type-based contextualized instance generator produces new instances based on given instances to enlarge the training set for better generalization. On three benchmark datasets, our model outperforms existing methods by significant margins.\footnote{Code can be found at \url{https://github.com/teapot123/Fine-Grained-Entity-Typing}.}
\end{abstract}




\maketitle

\section{Introduction}

Fine-grained entity typing (FET) aims to infer types of named entity mentions in specific contexts, which serves as the essential component for many downstream text mining applications, such as entity linking~\cite{Chen2020ImprovingEL}, knowledge completion~\cite{Dong2014KnowledgeVA}, text classification~\cite{Hu2019HeterogeneousGA}, named entity recognition~\cite{Huang2020FewShotNE}, question answering~\cite{Lin2012NoNP,Fader2014OpenQA}, \etc  
FET typically comes with a label hierarchy including both coarse-grained and fine-grained types, thus it is labor-intensive and time-consuming to annotate a large corpus for training fully-supervised models.
To mitigate the annotation burden for FET, some studies~\cite{Ren2016LabelNR, Chen2020HierarchicalET, Jin2019FineGrainedET, Dai2019ImprovingFE} explore distantly-supervised FET that denoises the pseudo labels automatically inferred from external knowledge bases; others~\cite{Ma2016LabelEF, Chen2021AnES} explore zero-shot FET that combines information from label hierarchy~\cite{Zhang2020MZETMA} and Wikipedia label descriptions~\cite{Zhou2018ZeroShotOE, Obeidat2019DescriptionBasedZF}. Our work focuses on the few-shot learning scenario, where a few annotated mentions within a particular context are given for each entity type in the label hierarchy, without requiring manual descriptions for labels.

% To deal with the challenge of limited training data, we leverage the strong generalization ability of
With the rapid development of pre-trained language models (PLMs) such as BERT~\cite{devlin2019bert} and RoBERTa~\cite{liu2019roberta} which are pre-trained on extremely large corpora with the masked language modeling (MLM) objective, PLMs have demonstrated remarkable text representation power and have been widely used as the backbone model for many downstream tasks, including entity typing.

The standard deployment of PLMs for entity typing is to fine-tune a PLM encoder together with a linear classifying layer which maps the PLM contextualized representations to the output entity types.
Such an approach is effective for fully supervised applications, but can easily lead to overfitting in few-shot scenarios~\cite{NNN2021PromptLearningFF} due to the random initialization of the linear layer---when the number of training samples is significantly smaller than the newly-introduced parameters, as is the typical scenario in the few-shot setting, training neural models becomes inherently unstable.

Recently, some studies~\cite{Petroni2019LanguageMA} show that without any further training of PLMs, factual knowledge can be probed from PLMs via natural language prompts that query the PLM's MLM head for masked word prediction. GPT-3~\cite{Brown2020LanguageMA} also found that prompts can guide the model to generate answers for specific tasks, which is especially effective on few-shot tasks. Inspired by these observations, prompt-based tuning has been widely developed for many text applications~\cite{Han2021PTRPT, Chen2021KnowPromptKP, Chen2021LightNERAL} and has generally surpassed vanilla fine-tuning in low-data regimes. 
There have also been explorations on automatic prompt searching~\cite{Jiang2020HowCW, Shin2020ElicitingKF} that aims to discover the optimal prompt templates for specific tasks.

Prompt-based tuning for few-shot FET has been explored very recently~\cite{NNN2021PromptLearningFF}:
A prompt template is concatenated with the original context containing the entity mention, and a PLM is used to predict the masked token in the template. Then a verbalizer maps the probability distribution over the vocabulary to a distribution over the entire label set. The verbalizer is typically constructed by extracting semantically relevant words to each label name from external knowledge bases. 

Despite the effectiveness of the proposed prompt-based tuning pipeline in~\cite{NNN2021PromptLearningFF}, there are several notable limitations: (1) The verbalizer construction method is only effective when the type label names have concrete and unambiguous meanings (\eg ``Person'' or ``Organization/Company''), but cannot be applied to label names that have composite semantics (\eg ``Location/GPE'' refers to geopolitical locations including countries, states/provinces and cities) or label names that share similar semantics (\eg ``Organization/Sports Team'' and ``Organization/Sports League). (2) The current prompt-based tuning approach has utilized the \textit{representation power} of PLMs to predict entity types based on instance representations, but have not fully explored the \textit{generation power} of PLMs acquired through extensive general-domain pre-training.

In this paper, we propose a novel framework for few-shot FET consisting of two modules: (1) An entity type label interpretation module automatically learns to relate the type labels to tokens in the vocabulary by jointly leveraging the given few-shot instances and the label hierarchy; (2) A type-based contextualized instance generator produces new instances based on few-shot samples and their types. The newly generated instances are added back to the training set with smoothed labels so that the model can be trained with more instances for better generalization. We demonstrate the effectiveness of our method by conducting experiments on three benchmark datasets where our model outperforms previous methods by large margins.
% We also conduct case studies and show the quality of new instances generated by two modules respectively. 
Our method is named \model, which stands for \textbf{A}utomatic \textbf{L}abel \textbf{I}nterpretation and \textbf{G}enerating \textbf{N}ew \textbf{I}nstance for \textbf{E}ntity typing.

The contribution of this paper are as follows:
\begin{itemize}[leftmargin=*]
\item We propose an entity type label interpretation module to capture the relation between type labels and tokens in the vocabulary leveraging both few-shot instances and the label hierarchy.
\item We propose a type-based contextualized instance generator to generate new instances based on few-shot instances and their types, which enlarges the training set for better generalization. 
\item On three benchmark datasets, our method outperforms existing methods by large margins.
\end{itemize}

\section{Preliminaries}
\begin{figure*}[!t]
\centering
\includegraphics[width=1\textwidth]{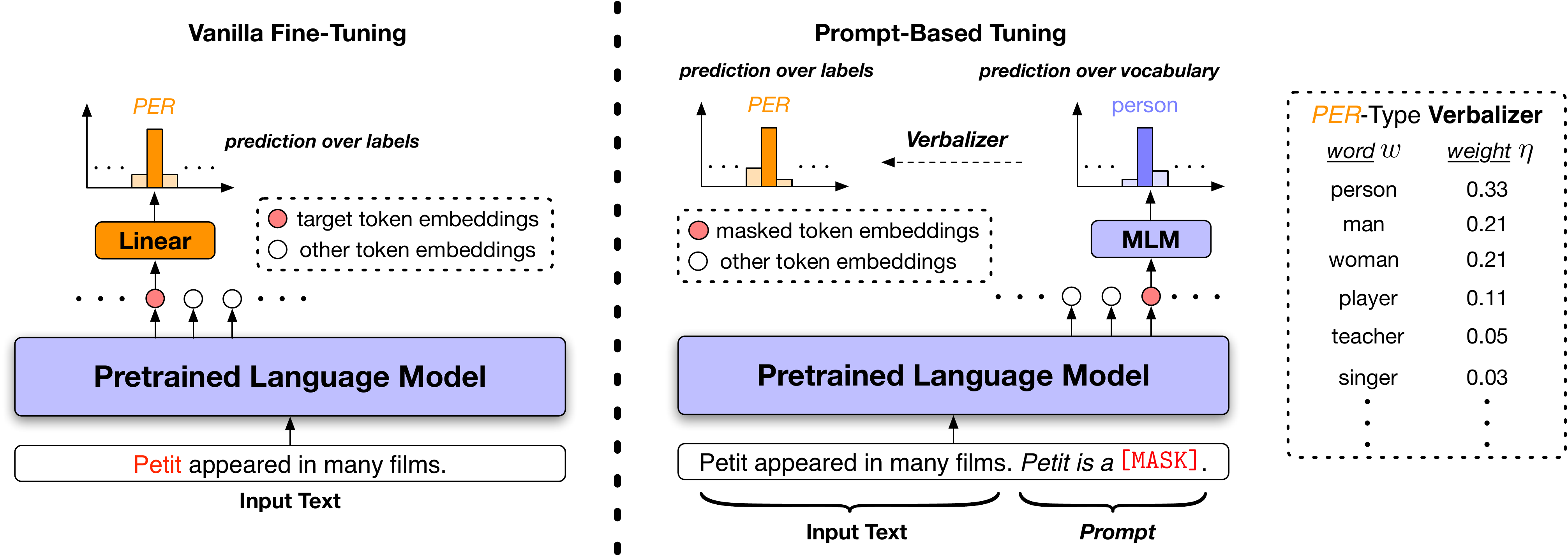}
\caption{(Left): Vanilla fine-tuning on entity typing task uses a linear classifier randomly initialized to project the contextualized representation into the label space. Under few-shot setting, the number of parameters in the linear classifier may far surpass the number of training samples, easily resulting in overfitting. (Right): Prompt-based tuning on entity typing task converts the entity type classification into a masked token prediction problem to reuse the extensively trained Masked Language Model (MLM) head in the pre-training stage without introducing new parameters, thus effectively transfers the prior pre-training knowledge into entity typing task. A verbalizer is then used to map the MLM output prediction over vocabulary to the final prediction over labels.}
\label{fig:prelim}
\vspace{-1em}
\end{figure*}

In this section, we first provide the definition of the few-shot entity typing task, and then briefly introduce vanilla fine-tuning and prompt-based tuning for entity typing which are illustrated in Fig.~\ref{fig:prelim}.

\noindent \textbf{Problem Formulation.}
Few-shot entity typing aims to determine the type of an entity based on its context. The input is a text sequence of length $T$, $\xv=\big\{ t_1, t_2, ..., \mv, ..., t_T \big\}$, where $\mv = \big\{ t_i, ..., t_j \big\}$ is an entity mention consisting of $(j-i+1)$ tokens, 
and the output is an entity type label $ y  \in \Ycal$ indicating the entity type of $\mv$ from a pre-defined set of entity types. The label set $\Ycal$ typically forms a hierarchical structure with both coarse and fine-grained types. The entity typing dataset is denoted with parallel triplets of context, mention, and type, as $\Dcal = \{ (\xv_i, \mv_i, y_i) \}_{i=1}^{|\Ycal|\times K} $ where $K$ is the number of example instances per type. In the few-shot learning setting, $K$ is typically small (\eg $K=5$). 

\noindent \textbf{Vanilla Fine-Tuning.}
Using a PLM such as BERT~\cite{devlin2019bert} as the encoder $\thetav_0$, the vanilla fine-tuning approach extracts the contextualized representation of each token $ \hv = f_{\thetav_0}(\xv)$ . A linear classifier (\ie a linear layer with parameter $\thetav_1 = \{\Wmat_1, \bv_1\} $  followed by a Softmax layer) is used to project the contextualized representation $\hv$ into the label space $f_{\thetav_1}(\hv) = \text{Softmax}(\Wmat_1 \hv + \bv_1)$.
Typically, the first token in the entity mention $\mv=\big\{ \tv_i, ..., \tv_j \big\}$ is used to compute the final prediction probability over the label space:
\begin{equation*}\label{eq:naiveft}
    p(y | \xv)=\text{Softmax}(\Wmat_1\cdot f_{\thetav_0}(\hv_i) + \bv_1)
\end{equation*}
The end-to-end learning objective for vanilla fine-tuning based entity typing can be formulated via function composition $f_{\thetav_1} \circ f_{\thetav_0}(\xv)$. The trainable parameters $\thetav = \{ \thetav_0, \thetav_1 \}$ are optimized by minimizing the cross-entropy loss:
\begin{equation}\label{eq:ce}
    \mathcal{L}_\text{ce} = -\sum_{i=1}^N \log  p(y_i |\xv_i) , 
\end{equation}
where $y_i$ is the gold label of $\xv_i$.
% and the KL divergence between two distributions is $\text{KL}(p || q)= \E_p \log (p/q) $.

\noindent \textbf{Prompt-Based Tuning.}
In vanilla fine-tuning, the parameters of the linear classifier $\thetav_1 = \{\Wmat_1, \bv_1\} $ are randomly initialized. Under the few-shot setting, the number of these newly-introduced parameters may far surpass the number of training samples, easily resulting in overfitting to the small training set. 
Prompt-based tuning, on the other hand, converts the entity type classification problem into a masked token prediction problem to reuse the extensively trained Masked Language Model (MLM) head in the pre-training stage without introducing any new parameters. 
The MLM head is used to predict what words can replace the randomly masked tokens in a text sequence. For a \mask token in the sequence, the MLM head takes its contextualized representation $\hv$ and outputs a probability distribution over the entire vocabulary $V$, which indicates the likelihood of a word $w$ appearing at the masked position:
\begin{equation}
\label{eq:mlm}
    p(w|\hv)=\text{Softmax}(\Emat\sigma(\Wmat_2 \hv +\bv_2)) 
\end{equation}
where $\Emat\in\R^{|V|\times h}$ is the embedding matrix; $\sigma(\cdot)$ is the activation function; $\Wmat_2\in\R^{h\times h}$ and $\bv_2\in\R^{h}$ are all pre-trained with the MLM objective. Since the parameters in both the MLM head $\thetav_2 = \{\Emat, \Wmat_2, \bv_2\}$ and the encoder $\thetav_0$ have been pretrained, prompt-based tuning effectively transfer the prior pretraining knowledge to the entity typing task for more stable and better performance.

% avoiding the usage of the classification layer $\thetav_1 = \{\Wmat, \bv\} $ in fine-tuning.
% so that instead of updating the randomly initialized parameters $\thetav_1$, the parameters $\thetav_0$ and $\thetav_2$ used by prompt-based tuning are already pre-trained in the PLM training objective. Intuitively, this stabilizes the down-stream task training in early epochs by leveraging prior knowledge via the MLM head.
\noindent \textbf{Template and Verbalizer for Prompt-Based Tuning.}
A prompt is composed of two critical components: A template $T_{c}$ that forms a cloze-style ``fill-in-the-blank'' sentence to perform MLM head prediction and a verbalizer $\Vcal$ that relates predicted words to entity labels. 
The use of prompt-based tuning on the entity typing task was first explored by~\cite{NNN2021PromptLearningFF}. For a sentence $\xv$ with an entity mention $\mv$, a valid template can be as follows:
\begin{align}
T_{c}(\xv, \mv) = \xv\text{. } \mv \text{ is a \mask.}\nonumber
% T_{c_2}(\xv, \mv) &= \xv\text{. } \mv \text{ is a type of [MASK].}\nonumber\\
% T_{c_3}(\xv, \mv) &= \xv\text{. In this sentence, } \mv \text{ is a [MASK].}\nonumber
\end{align}
The MLM head will output the likelihood of each word $w$ in the vocabulary $V$ appearing at the \mask position.
% Additionally, soft prompts are proposed to replace discrete tokens with continuous vectors, to avoid searching for optimal discrete templates. Apart from the above hard templates, the following soft template can be created:
% $$T_{c_4}(\xv, \mv) = \xv\ \pv_0\ \mv\ \pv_1\ \pv_2\ ... \pv_{l-1}\ \text{[MASK] } \pv_l$$
% where each continuous vector $\pv_i$ can be initialized and updated using the embedding of the original word in the hard templates. These soft vectors can be thought of as a grouping of semantically related words.
Then the verbalizer associates output words with each label. For instance, words like \{business, corporation, subsidiary, firm, ...\} may be associated with the entity label ``Organization/Company'', so that the output probability for these words will be added up as the prediction for their associated type label. 
Suppose the verbalizer for category $y$ is $\Vcal_y=\{w_1, ..., w_m\}$, then the final label prediction probability of $\xv$ is calculated as follows:
$$p(y|\xv)=\frac{1}{|\Vcal_y|}\sum_{w_i\in \Vcal_y} \eta_i
p(w_i|\hv)$$
% p(\text{[MASK]}=w_i|T(\xv))$$
where $\eta_i$ is a weight indicating the importance of word $w_i$ and the final training objective is the same as Eq.~\eqref{eq:ce}.
The quality of verbalizer is critical to the final performance, and previous prompt-based tuning methods either manually select type-related words~\cite{Schick2021ExploitingCF} or leverage external knowledge graph~\cite{NNN2021PromptLearningFF, Chen2021KnowPromptKP} to automatically extract related nouns to the label names.

\section{Method}

\begin{figure*}[t]
\centering
\includegraphics[width=1.0\textwidth]{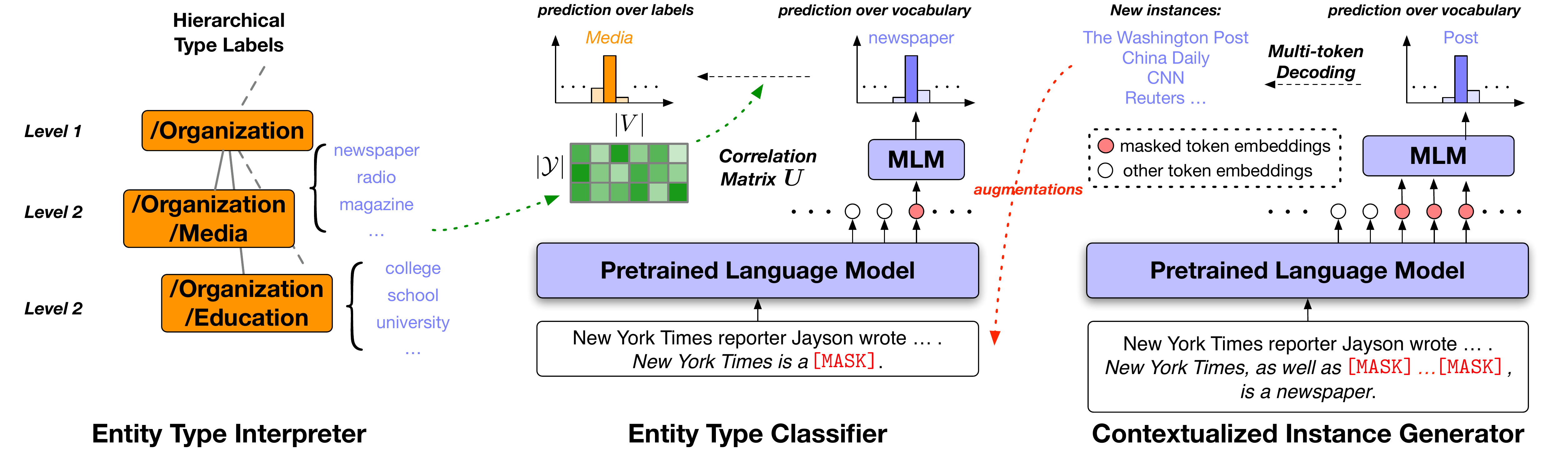}
\caption{Overall framework of \model. (Left): With a given type label hierarchy, an entity type interpretation module relates all the words in the vocabulary with the label hierarchy by a correlation matrix, so that the top-related words of ``/Organization/Media'' are ``newspaper'', ``radio'' and ``magazine''. (Middle): With an entity mention and its context from the training set, an entity typing classifier uses a template and an MLM head to predict word probability at the \mask position, and then maps the word distribution to type probability using the correlation matrix. The word ``newspaper'' has a high relevance with the label ``/Organization/Media''. (Right): After the entity typing classifier predicts a type for an entity mention in the few-shot samples, a type-based contextualized instance generator uses that entity mention and the predicted type to construct a template for new instance generation. The newly generated instances for ``New York Times'' are: ``The Washington Post'', ``China Daily'', \etc They are added back to the training set to improve the generalization of the entity typing classifier.}
\label{fig:framework}
\end{figure*}

Despite the effectiveness of current prompt-based tuning pipeline, there are two notable limitations: 
(1) 
% he MLM head is trained to predict the original token at the \mask position, thus the prediction space is the whole vocabulary $V$ instead of the smaller label set $\Ycal$. 
Current automatic verbalizer construction methods extract type representative words to be mapped to the label set based on general knowledge (\eg knowledge bases), but they are only effective when the type label names have concrete and unambiguous meanings (\eg ``Organization/Company''). 
When the label names are too abstract or have composite meanings (\eg ``Location/GPE'' refers to geopolitical locations including countries and cities), however, suitable type indicative words cannot be easily generated only via external knowledge.
Meanwhile, it is challenging to distinguish semantically similar types (\eg ``Organization/Sports Team'' and ``Organization/Sports League'') well for robust fine-grained type classification.
(2) The current prompt-based fine-tuning approaches have not fully utilized the power of PLMs: PLMs are commonly used as the \emph{representation models} that predict entity types based on entity instance representations. 
On the other hand, the \emph{generation power} of PLMs acquired through extensive general-domain pretraining can be exploited to  generate new entity instances that do not exist in the few-shot training samples so that the entity typing model can be trained with more instances for better generalization.

To address the above limitations, we propose two modules for prompt-based few-shot entity typing: 
(1) An \emph{entity type label interpretation module} automatically learns to relate the hierarchical entity type labels to tokens in the vocabulary by jointly leveraging the given few-shot instances and the label hierarchy; 
(2) A \emph{type-based contextualized instance generator} generates new instances based on few-shot instances and their types, providing the type classifier with more training samples.
Our overall framework is illustrated in Fig.~\ref{fig:framework}.
% We will introduce the two modules in detail in the following sections.

\subsection{Hierarchical Entity Label Interpretation}
Interpreting label semantics with type-indicative words is crucial for prompt-based tuning in entity typing, as the final predictions are obtained by mapping PLMs' output words to entity labels.
A recent study~\cite{NNN2021PromptLearningFF} proposes to use external knowledge bases (\eg Related Words\footnote{https://relatedwords.org}) to find type-related words for constructing the verbalizers. 
For example, the related words retrieved for label ``Company'' are \textit{business, corporation, subsidiary, firm, }\etc
While such an approach works well for common entity types, it suffers from two notable limitations: (1) It merely relies on external knowledge without considering the specific target corpus information which is essential for interpreting complicated or abstract labels. (2) It does not utilize the label correlations in the type hierarchy which may help the distinction and interpretation of fine-grained label semantics.
In the following, we introduce how to jointly leverage few-shot samples and the entity label hierarchy to learn a label interpretation module.
% Fine-grained entity typing datasets usually come with a label hierarchy where entity mentions can belong to a middle-level coarse type or a fine-grained type on the leaf node. 
% Prompt-based method leverages MLM head for probing the factual knowledge in PLMs. Despite its effectiveness, the output space of MLM head (the whole vocabulary $V$) is much larger than the label set/hierarhcy of entity typing datasets.
% Directly enforcing the MLM head to only predict the label names for [MASK] tokens in templates will harm the distribution of the embedding matrix $E$ and does not result in good performance.
% Understanding and distinguishing the label semantics is crucial for correctly mapping possible MLM head outputs to the type label set.
% When a human try to understand the semantics of a particular label, he/she most likely will check the entity mentions in the few-shot examples of that label as well as its relation with other labels in the hierarchy, to determine which more fine-grained types belong to that label. 
% Therefore, our label decomposition module utilizes both sources of information to fulfill the word-to-type mapping.

\noindent \textbf{Learning Word-Type Correlation Matrix.}
We consider a correlation matrix $\Umat\in\R^{|\Ycal| \times |V|}$ that characterizes the correlation between each word $w \in V$ and each type label $y \in \Ycal$. Specifically, each element $u_{y,w}$ (the $y$-th row and $w$-th column) in $\Umat$ represents a learnable correlation score between $w$ and $y$.
% , so that each element represents $P(y|\wv)$. 
% After we get the prediction probability of a \mask token $P(\texttt{[MASK]}=w)$ for $w \in V$, we can then obtain the output probability on the label space $y_i\in \Ycal$ by the type relevance of each word $w$ to each type $y$:
The correlation matrix maps the token predictions of PLMs (\ie $p(w|\hv)$ in Eq.~\eqref{eq:mlm}) to the final entity label predictions $p(y|\hv)$:
\begin{align}\label{eq:decomp}
    p(y|\hv) &= \sum_{w\in V} p(y|w) p(w|\hv),
    % &= \text{Softmax}(\sum_{w\in V}\uv_{y_i,w}\Emat\sigma(\Wmat_2 \hv +\bv_2))
\end{align}
where 
$$
p(y|w) = \frac{\exp(u_{y,w})}{\sum_{y'}\exp(u_{y',w})}.
$$
When Eq.~\eqref{eq:decomp} is trained via cross entropy loss on few-shot samples, the correlation matrix $\Umat$ will be updated to automatically reflect the corpus-specific word-to-type correlation.
For example, consider an input text with ``Ukraine'' as a ``GPE''-type entity, the output of the MLM head at the \mask position is very likely to give high probability to words like ``country'' or ``nation'', therefore the corresponding elements $u_{y,w}$ with $y$ being ``GPE'' and $w$ being ``country'' or ``nation'' will increase during training.

To further grant $\Umat$ a good initialization, we assign a higher initial value to those words that are one of the label names of an entity type. Specifically, the elements in $\Umat$ are initialized to be:
\begin{equation}\label{eq:init_U}
\large
u_{y, w} = \begin{cases}
\frac{1-\alpha}{|\yv|}+\frac{\alpha}{|V|-|\yv|} & w \in \yv \\
\frac{\alpha}{|V|-|\yv|} & w \notin \yv
\end{cases},
\end{equation}
where $\yv$ denotes the label name set of label $y$; $\alpha$ is a hyperparameter controlling the initial bias to label name match (\eg when $\alpha$ is close to $0$, words that do not match the label names will have a near-zero initialization).
% so that words in a label name are more related to that label than other words, which are equally weighted at the initial point. 
% For example, the word \textit{sports} will contribute equally to both ``Sports Team'' and ``Sports League'' types.
% Then the elements will be updated by training on the few-shot examples to learn the relevance between words and types, so that more fine-grained type words will be related to their corresponding type labels.
% We can simply use the cross entropy loss on few-shot examples to learn the mapping between words and types:
% \begin{equation}\label{eq:ce_loss}
%     \mathcal{L}_{\text{ce}} =  -\sum_{i=1}^N \log  p(y=y_i |\xv_i), 
% \end{equation}

\smallskip
\noindent \textbf{Regularization with Hierarchical Information.}~
Apart from the information from few-shot examples, the labels of an FET dataset typically forms a hierarchy that specifies the \emph{inclusive} relationship between parent-child labels and the \emph{exclusive} relationship between sibling labels. 
Therefore, we can further regularize the correlation matrix $\Umat$ with the hierarchical information. Following~\cite{Meng2020HierarchicalTM} that learns a spherical tree embedding given a hierarchical structure, we learn the correlation matrix by considering the relationship of nodes on the hierarchy.

To model the semantic inclusiveness between parent-child label pairs, we impose an inclusive loss to relate the parent label with its child labels by minimizing their cosine distance:
\begin{align}\label{eq:inc_loss}
&\mathcal{L}_{\text{inc}} = \sum_{y\in \Ycal} \left( 1- \cos\left(\uv_{y}, \uv_{\text{Parent}(y)}\right) \right),
\end{align}
where $\uv_{y}$ refers to the $y$-th row in $\Umat$, and Parent$(y)$ means the parent node of $y$ in the label hierarchy. Eq.~\eqref{eq:inc_loss} encourages each label to have a similar correlation score distribution to all its children labels (\eg the ``Organization'' label is the broader type of its children labels ``Company'', ``Sports Team'', and ``Sports League''; ``Organization'' will be thus regularized to share all the relevant words to any of its children labels). 

Meanwhile, we also aim to distinguish sibling type categories like ``Sports Team'' and ``Sports League'' for better distinctive power at finer-grained levels. 
Therefore, we apply an exclusive loss that minimizes the cosine similarity between each pair of sibling labels sharing the same parent node in the hierarchy:
\begin{equation}\label{eq:exc_loss}
    \mathcal{L}_{\text{exc}} = \sum_{\substack{y_i, y_j\in \Ycal\\y_i\neq y_j}} \cos(\uv_{y_i}, \uv_{y_j}), \  \text{Parent}(y_i)=\text{Parent}(y_j).
\end{equation}

Finally, the overall learning objective for the label interpretation module is:
\begin{equation}\label{eq:dec_loss}
    \Lcal = \Lcal_{\text{ce}} + \lambda \Lcal_{\text{exc}} + \lambda \Lcal_{\text{inc}},
\end{equation}
where $\Lcal_{\text{ce}}$ is the cross entropy loss in Eq.~\eqref{eq:ce}; $\lambda$ is the weight for regularization.

\subsection{Type-Based Contextualized Instance Generator}
Previous prompt-based methods mainly utilize the \emph{representation power} of PLMs: By learning contextualized representations of the input texts, PLMs are able to accurately predict the \mask tokens which are then mapped to the final entity typing predictions.
In fact, in a token-level task like FET, not only can PLMs infer the type of an entity mention, but it also may generate new instances of specific entity types based on pretraining knowledge.
In text classification studies~\cite{Meng2020TextCU}, topic-related words are generated through pre-trained MLM for self-training.
Similarly, we explore the \emph{generation power} of PLMs to construct new instances using existing instances and their types as examples. 
For example, given an entity ``Buffalo'' and a sentence ``At both Delaware State and \textbf{Buffalo}, ..., as both schools transitioned from lower levels of NCAA hierarchy'', the MLM head in the typing classifier predicts the entity ``Buffalo'' as a \textit{university}.
Based on the predicted type, our goal is to generate new instances of the same type, such as other universities like ``Duke'' and ``Dartmouth''.
These newly found instances can then be added back to the training set for better generalization of the typing classifier.

To fulfill the goal of generating same-type instances from an example entity $\mv$, we need to first use the typing classifier to predict the fine-grained type of $\mv$, denoted as $t$, and then create a generative template $T_g(\mv, t)$ to generate new instances:
$$T_g(\mv, t) = \mv \text{, as well as \mask, is a } t\text{.}$$
Using the above template, the MLM head will try to predict the instances with the same type $t$ as entity mention $\mv$. 

\smallskip
\noindent \textbf{Multi-Token Instance Generation.}~
The above proposed template can already generate single-token instances effectively, but will obviously miss a large amount of multi-token instances. 
For example, when we want to generate new instances for 
% the sentence ``The book also appeared on the 
\textbf{New York Times}, 
% Bestseller List'', 
the above method will only generate single-token newspaper names like ``Reuters'', ``CNN'', \etc 
Therefore, to generate multiple-token instances, we inject a sequence of \mask tokens into the template $T_g(\mv, t)$. 
\begin{equation}
\label{eq:temp_multi}
T_g(\mv, t) =  \mv \text{, as well as \mask \dots \mask, is a } t\text{.}    
\end{equation}
% In this case, since the original mention \textbf{New York Times} is tokenized into three subwords, we also include three \mask tokens to obtain same-type instances. 
Such a multiple-token instance generation process is necessary even for single-word entities, since they can be potentially sliced into multiple subwords by PLM tokenization (\eg, ``Chanel'' will be sliced into ``Chan'' and ``\#\#el'' in BERT tokenization), and using multiple \mask positions for prediction will give chance to the generation of less frequent words that consist of multiple subwords. 
% The original BERT pre-training~\cite{devlin2019bert} randomly samples 15\% tokens in a sentence, and replace 80\% with [MASK] tokens, thus there is no limitation in the amount of [MASK] token per sentence.
Multi-token decoding is commonly carried out in autoregressive language models~\cite{Radford2019LanguageMA} as sequential left-to-right steps. 
However, it becomes less straightforward for autoencoder models like BERT that uses bidirectional contexts for prediction:
While BERT can predict multiple \mask tokens in a sequence simultaneously by greedily picking the most likely token at each individual \mask position, these predictions are made independently from each other and the final results may not make sense as a whole.
% Since our goal is to find real entity names, we cannot directly take the highest predicted word individually for all \mask positions, which do not usually result in a meaningful collocation.
Therefore, we propose to fill in one blank at each step by selecting the word with the highest score at that position, and recursively predict the other blanks conditioned on the already filled blanks. Later we will score the multiple combinations of words and select the top-$M$ combinations as the new instances.
For example, using $T_g(\mv, t)$ and decoding the masked tokens from left to right, we can generate a new instance that has the same type with \textbf{New York Times} in the following steps:
\begin{figure}[!h]
    \centering
    \includegraphics[width=1.0\columnwidth]{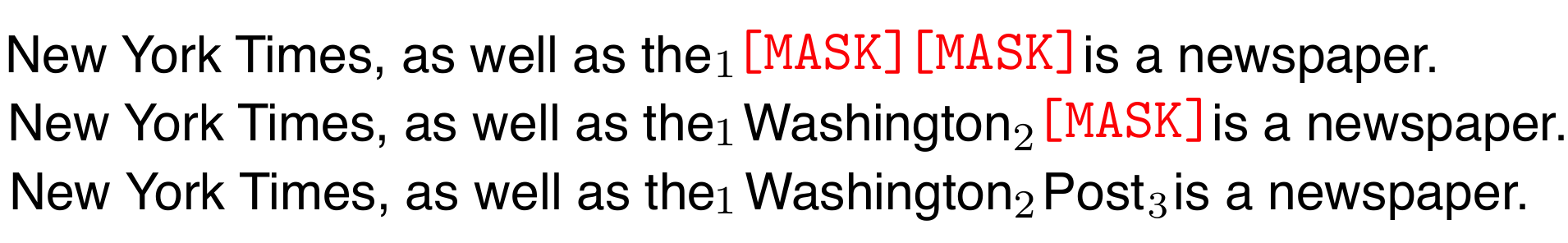}
\end{figure}
\\Since there is no ground truth for the length of the new instances, we iterate from $1$ to $l$ \mask tokens, where $l$ is the length of the example entity. We then rank all generated new instances $\widetilde{\mv}$ with pseudo log likelihood~\cite{Jiang2020XFACTRMF} to select top new instances:
$$
\text{Score}(\widetilde{\mv})=\sum_{i=1}^{|\widetilde{\mv}|}\log(s_i),
$$
where $s_i$ is the conditional probability of predicting the $i$-th sub-word based on previous $i-1$ generated sub-words.

\smallskip
\noindent \textbf{Contextualized Instance Generation.}~
Apart from probing the knowledge in PLMs to generate new instances, sometimes entity instances of the same type may appear in parallel. 
For example, the context ``..., while Dan Oster, Anjelah Johnson, and Daheli Hall were hired as feature players'' include multiple \textit{players}/\textit{actors} such as ``Dan Oster'', ``Anjelah Johnson'' and ``Daheli Hall''.
In the few-shot learning setting where a limited number of training examples at instance-level are provided, it is likely that the above sentence, if appearing in the training set, only contains one annotated entity, and other entities in the same sentence will not appear in the training set.
To find potential parallel entity instances in the sentence, we extend the template in Eq.~\eqref{eq:temp_multi} by adding the original context $\xv$ to form our final instance generation template:
$$T_g(\xv, \mv, t) = \xv\text{. } \mv \text{, as well as \mask} \dots \text{\mask}\text{, is a } t\text{.}$$
Using the above template, both in-context parallel instances and out-of-context new instances based on pre-training knowledge can be generated through the instance generator.

\smallskip
\noindent \textbf{Adding New Instances for Training.}~
For each given entity mention $\mv$, we can generate $M$ new instances after the training finishes half of the total epochs when the type words output by the MLM head become stable.
The type-based new instances are then added back to augment the training set by applying the same classification template. 
Since there could be noise in generating new instances, we do not impose a hard label on a new instance $\widetilde{\mv}$ to train the model.
Instead, to improve robustness to label noise~\cite{Lukasik2020DoesLS}, we use a soft distribution over $\Ycal$ as its pseudo label $\widetilde{\yv}$ by smoothing the label $y_i$ of the original entity $\mv$:
\begin{equation}\label{eq:label_smooth}
\widetilde{\yv} = \begin{cases}
1-\epsilon+\frac{\epsilon}{|\Ycal|} & y=y_i \\
\frac{\epsilon}{|\Ycal|} & y\neq y_i
\end{cases},
\end{equation}
where $\epsilon$ is the smoothing parameter.

We use the KL divergence loss to train the model on newly generated instances:
\begin{equation}\label{eq:kl_loss}
    \mathcal{L}_{\text{new}} = \sum_{i=1}^N \sum_{k=1}^M \text{KL}(\widetilde{\yv}_{i,k}||p(y |\widetilde{\mv}_{i,k})), 
\end{equation}
where $\widetilde{\mv}_{i,k}$ is the $k$-th instance generated from the $i$-th entity mention, and $\widetilde{\yv}_{i,k}$ is its corresponding label.

\subsection{Overall Training Objective}
The overall training objective is the combination of training the label interpretation module and using new generated instances as the augmentations:
\begin{equation}\label{eq:dec_loss}
    \Lcal_\text{total} = \Lcal_{\text{ce}} + \lambda \Lcal_{\text{exc}} + \lambda \Lcal_{\text{inc}} + \beta\lambda_n \Lcal_{\text{new}}
\end{equation}
During the first half of training epochs, $\beta$ is set to $0$ to avoid generating false types for existing entities.
After half of the training epochs finish, the model becomes more stable on predicting types, and we gradually increase $\beta$ as the epoch grows:
$\beta=\frac{2t-T}{T}$, where $T$ is the total number of epochs and $t$ is the current epoch number.

% to be modified:
% dataset(x, m, c)
% label decomposition (mismatch between predictions and labels)

\section{Experiment}
In this section, we conduct experiments to demonstrate the effectiveness of our method. We introduce the three benchmark datasets on FET in Sec~\ref{sec:data} and details of experiment settings in Sec~\ref{sec:exp_set}. Then we report the main quantitative results in Sec~\ref{sec:results} and analyze some case studies in Sec~\ref{sec:case}.

% \begin{table}[ht]
% \centering
% \caption{
% Statistics of each few-shot sampled dataset with detailed training/test split of entity mentions and the depth/number of entity types. We sampled 10 times for each dataset.
% }
% \begin{tabular}{*{6}{c}}
% \toprule
% \textbf{Dataset} & \textbf{\# Train} & \textbf{\# Test} & \textbf{\# Types} & \textbf{Depth}\\
% \midrule
% OntoNotes & 105  & 8,963 & 21 & 3 \\
% BBN & 125 & 12,824 & 25 & 2\\
% Few-NERD & 330  & 96,902 & 66 & 2 \\
% \bottomrule
% \end{tabular}
% % \vspace{-0.5em}
% \vspace{-1em}
% \label{tab:dataset}
% \end{table}

\begin{table*}[!t]
\caption{
Results on three entity typing benchmark datasets. For 5-shot setting, we report the average performance over 5 different sets of randomly-sampled few-shot examples.
}\label{tab:main}
\vspace{-1em}
\centering
\begin{tabular}{l*{9}{c}}
\toprule
\multirow{2}{*}{\textbf{Method}} 
& \multicolumn{3}{c}{\textbf{OntoNotes}}
& \multicolumn{3}{c}{\textbf{BBN}}
& \multicolumn{3}{c}{\textbf{Few-NERD}}\\
& (Acc.) & (Micro-F1) & (Macro-F1) & (Acc.) & (Micro-F1) & (Macro-F1) & (Acc.) & (Micro-F1) & (Macro-F1)  \\
\midrule
\multicolumn{10}{l}{\textbf{5-Shot Setting}}  \\ 
\midrule
Fine-tuning & 28.60 & 50.70& 51.60& 51.03 & 60.03 & 58.22 & 36.09 & 48.56 & 48.56 \\
Prompt-based MLM & 32.62 & 60.97 & 61.82 & 67.00 & 75.23 & 73.55 & 44.69 & 59.24 & 59.24 \\
PLET & 48.57 & 70.63 & 75.43 & 71.23 & 79.22 & 78.93 & 56.94 & 68.81 & 68.81 \\
\model (- hierarchical reg.) & 52.74 & \textbf{77.55} & 79.72 & 72.15 & 80.35 & 80.40 & 59.01 & 70.91 & 70.91 \\
\model (- new instances) & 51.10 & 72.91 & 76.88 & 73.50 & 81.62 & 81.31 & 57.41 & 69.47 & 69.47 \\
\model  & \textbf{53.37} & 77.21 & \textbf{80.68} & \textbf{75.44} & \textbf{82.20} & \textbf{82.30} & \textbf{59.72} & \textbf{71.90} & \textbf{71.90}\\
\midrule
\multicolumn{9}{l}{\textbf{Fully Supervised Setting}}  \\ 
\midrule
Fine-tuning & 56.70 & 75.21 & 78.86 & 78.06 & 82.39 & 82.60 & 79.75 & 85.74 & 85.74 \\
Prompt-based MLM & 55.18 & 74.57 & 77.47 & 77.10 & 81.77 & 82.05 & 77.38 & 85.22 & 85.22 \\
\bottomrule
\end{tabular}
\end{table*}

\subsection{Datasets}\label{sec:data}
We use three fine-grained entity typing benchmark datasets.
\begin{itemize}[leftmargin=*]
\item \textbf{OntoNotes}. The OntoNotes dataset is derived from the OntoNotes corpus~\cite{Ontonotes} and 12,017 manual annotations were done by~\cite{Gillick2014ContextDependentFE} with 3 hierarchical layers of 89 fine-grained types. We follow the dataset split by~\cite{Shimaoka2017NeuralAF} that retains 8,963 non-pronominal annotations, where the training set is automatically extracted from Freebase API by~\cite{Ren2016LabelNR}. Since our paper focus on few-shot learning, we apply extra preprocessing by filtering out classes with less than 5 annotations in training, validation, and test set, resulting in a total of 21 classes left.
\item \textbf{BBN Pronoun Coreference and Entity Type
Corpus} (BBN). This dataset uses 2,311 Wall Street Journal articles and is annotated by~\cite{BBN} with 2 hierarchical layer of 46 types. We follow the split by~\cite{Ren2016LabelNR} and also filter out classes with less than 5 annotations in training, validation and test set, leading to a total of 25 classes.
\item \textbf{Few-NERD}. The Few-NERD~\cite{NNN2021FewNERDAF} dataset is a recently proposed large-scale manually annotated dataset with 2 hierarchical layers of 66 types. We follow~\cite{NNN2021PromptLearningFF} and uses the supervised setting of the dataset,
FEW-NERD (SUP), as well as the official split for few-shot example sampling. All 66 classes have more than 5 annotations in each of the dataset splits.
\end{itemize}
% The detailed statistics of the three datasets are listed in Table~\ref{tab:dataset}.

\subsection{Experiment Settings}\label{sec:exp_set}
\smallskip
\noindent \textbf{Few-Shot Sampling.}~
We conduct 5-shot learning on three datasets by sampling 5 instances for training and 5 instances for dev set in each run of experiment. We repeat experiments for each dataset for 5 times with different sampled sets and report the average result.

\smallskip
\noindent \textbf{Compared Methods and Ablations.}~
We include the results of vanilla fine-tuning, prompt-based MLM using hard templates in~\cite{NNN2021PromptLearningFF}, and the current state-of-the-art method \textbf{PLET}~\cite{NNN2021PromptLearningFF} on few-shot entity typing, which uses multiple hard and soft templates for prompt-based tuning, and automatically constructs verbalizers through external knowledge graph. 
We conduct ablation studies by removing the following two parts one at a time: (1) not using the hierarchical regularization in Eq.~\eqref{eq:exc_loss} and Eq.~\eqref{eq:inc_loss} (- hierarchical reg.); (2) not using the type-based contextual instance generator to generate new instances for training in Eq.~\eqref{eq:kl_loss} (- new instances).
We also compare with vanilla fine-tuning and prompt-based tuning in a fully supervised setting.

\smallskip
\noindent \textbf{Hyperparameter Settings.}~
We use the pre-trained RoBERTa-base model as the backbone transformer model (for \model and all baselines). For all three datasets: the max sequence length is set to be $128$; the batch size is $8$; the training epoch number is $30$; the hyperparameters $\alpha$ and $\epsilon$ are set to $0.1$; the instance generating number $M$ is set to $5$ per type; the training weights $\lambda$ and $\lambda_n$ are set to $1.0$. We use a learning rate of $1e-2$ for the correlation matrix $U$ and a gradient multiplication layer of $1e-7$ is applied to the bottom Transformer backbone so that the pre-trained weights stay very stable during the training.
% PLET and Prompt-based MLM use a learning rate of $5e-5$ as suggested by their paper~\cite{NNN2021PromptLearningFF}. Fine-tuning uses a learning rate of $1e-4$ as suggested by~\cite{} since an extra neural layer is applied.
We use Adam~\cite{Kingma2015AdamAM} as the optimizer with linear decaying schedule.
The model is run on NVIDIA GeForce GTX 1080 Ti GPU.

\smallskip
\noindent \textbf{Evaluation Metrics.}~
We apply the widely-used metrics from~\cite{Ling2012FineGrainedER} consisting of strict accuracy (Acc.), loose micro-F1 score (micro-F1), and loose macro-F1 score (macro-F1). The loose F1 scores tolerate partial correctness for type labels within the same branch but of different granularities.

\begin{table*}[ht]
\centering
\caption{Top generated instances based on predicted types of example entities.}\label{tab:new_inst}
\vspace{-0.5em}
\begin{tabular}{|c|c|c|}
\hline
\multicolumn{3}{|c|}{Generation from \textbf{single-token} entities}\\
\hline
Context \& \textbf{entity mention} & MLM predicted type & Generated new instances\\
\hline
\makecell{From Coeus and Phoebe came \textbf{Leto} and Asteria, who married \\Perses, producing Hekate, and from Cronus and Rhea came Hestia,\\ Demeter, Hera, Poseidon, Hades, and Zeus.
} & god 
& \makecell{Zeus, Hermes, Hades, Apollo, Athena,\\ Hera, Pluto, Prometheus, ...}\\
\hline
\makecell{Orsonwelles malus is a species of spider endemic to \textbf{Kauai}\\ in the Hawaiian Islands.} &
island & \makecell{Hawaii, Fiji, Samoa, Guam, Taiwan,\\ Honolulu, Japan, ...}\\
\hline
\makecell{At both Delaware State and \textbf{Buffalo}, Townsend was responsible for\\ leading the athletic department to achieve full NCAA Division I status,\\ as both schools transitioned from lower levels of NCAA hierarchy.} & university & \makecell{Delaware, Wilmington, Syracuse,\\Albany, Rutgers,\\Dartmouth, Duke, ...}\\
\hline
\makecell{A well-known philanthropist, Shaw donated billions of\\Hong Kong \textbf{dollars} to educational institutions in Hong Kong\\and mainland China.} & currency & \makecell{yuan, euros,\\sterling, yen,\\bitcoin, pounds, ...}\\
\hline
% \makecell{An aphorism was used and credited in a 2011\\ marketing campaign by the French cognac brand, \textbf{Courvoisier}.} & drink & \makecell{vodka, whisky,\\champagne, gin,\\bourbon, ...}\\
% \hline
% \makecell{The \textbf{NFL} would do this for all but one\\ Super Bowl after this until Super Bowl XXXI.} & league & \makecell{MLB, NCAA,\\NBA, NHL,\\CFL,FIFA, ...}\\
% \hline
% \makecell{Asheville Regional Airport is a focus city for \textbf{Allegiant Air}\\ who bases Airbus A320 family aircraft and crew at the\\ airport .} & airline &
% \makecell{Alaska Airlines, Canadian Airways, \\Frontier Airlines, Spirit Airlines,\\ Virgin America, Qatar Airways, ...}\\
% \hline
% \makecell{He is followed by fellow Mexican - American performer Pepe Aguilar\\ with three winning albums and by American singers Vikki Carr\\ and Linda Ronstadt, Mexican singers Luis Miguel and Joan Sebastian,\\ and bands La Mafia and \textbf{Los Lobos}, with two wins each.} & band & \makecell{La Mafia, The Clash,\\The Beatles, The Mob,\\Black Flag, ... }\\
% \hline
\multicolumn{3}{|c|}{Generation from \textbf{multi-token} entities}\\
\hline
Context \& \textbf{entity mention} & MLM predicted type & Generated new instances\\
% \hline
% \makecell{\textbf{OSF Saint Francis Medical Center} started Peoria's largest-ever\\private building expansion to build a new emergency room and\\a new Children's Hospital of Illinois; and Methodist\\Medical Center of Illinois and Pekin Hospital
% also expanded.} & hospital & \makecell{Children's Hospital of Illinois,\\Methodist Medical Center of Illinois,\\Children 's Medical Center,\\OSF Mercy Medical Center,\\Northwestern Memorial Hospital, ...}\\
\hline
\makecell{The album also included the song ``Vivir Lo Nuestro,''\\ a duet with \textbf{Marc Anthony}.} & singer & \makecell{Beyonce, Jennifer Lopez,\\Rihanna, Taylor Swift, \\ Lady Gaga, Michael Jackson, ...}\\
\hline
\makecell{The film was released on August 9, 1925, by \textbf{Universal Pictures}.} & company & \makecell{Warner Brothers, Paramount Pictures ,\\Columbia Pictures, Lucasfilm, \\ Hollywood Pictures, ...}\\
\hline
\makecell{Everland hosted 7.5 million guests in 2006, ranking it fourth\\in Asia behind the two \textbf{Tokyo Disney Resort} parks and\\Universal Studios Japan, while Lotte World attracted 5.5 million\\ guests to land in fifth place.} & park & \makecell{Lotte World, Universal Studios Japan,\\Shanghai Disney World ,\\Orlando Universal Studios, ...}\\
\hline
\makecell{Their daughter, Caroline Montgomery Marriott, was killed by the\\ \textbf{Spanish Flu} epidemic in 1918.} & disease & \makecell{Yellow Fever, bird flu,\\typhus, small pox,\\polio, ...}\\
\hline
\makecell{The disappearance of the Norse settlements probably resulted\\ from a combination of the Little Ice Age 's cooling temperatures,\\abandonment after the \textbf{Black Plague} and political turmoils.} & disaster & \makecell{climate change, Ice Age,\\global warming, Norman Conquest,\\cannibalism, natural disasters, ...}\\
\hline
\makecell{The site of Drake's landing as officially\\recognised by the \textbf{U.S. Department of the Interior}\\and other agencies is Drake's Cove.} & government agency & \makecell{the Department of Homeland Security,\\the Bureau of Land Management,\\the Federal Bureau of Investigation,\\the United States Forest Service,\\the National Institutes of Health, ...}\\
\hline
\makecell{Pikmin also make a cameo during the process\\ of transferring downloadable content from a \textbf{Nintendo DSi}\\ to a 3DS, with various types of Pikmin carrying the data over.} & handheld & \makecell{3DS, 2DS,\\Wii U, Nintendo Switch,\\the PSP, PlayStation Vita, ...}\\
\hline
\end{tabular}
\end{table*}

\subsection{Quantitative Evaluation}\label{sec:results}
% \smallskip
% \noindent \textbf{Main Results.}~
Table~\ref{tab:main} presents the performance of all methods on three benchmark datasets. Overall, prompt-based results have higher performance than vanilla fine-tuning in few-shot settings, showing the prompts are better at inducing factual knowledge from PLMs. In fully supervised settings, however, fine-tuning performs a little better than prompt-based MLM, which is also suggested by previous studies~\cite{Gu2021PPTPP}, because vanilla fine-tuning has an extra linear layer to learn more features from the full training set. 
Specifically, \model achieves the best among all prompt-based methods on all three datasets, and is mostly better than the two ablations, demonstrating that the two proposed modules can effectively find related words and generate new instances for each label.
We also notice that \model can perform on par with fully supervised setting on \textbf{OntoNotes} and \textbf{BBN}, but cannot on \textbf{Few-NERD}. This is because the training set of \textbf{OntoNotes} and \textbf{BBN} are automatically inferred from external knowledge bases, and can contain much noise, while the training set of \textbf{Few-NERD} is totally labeled by human. 
Details of hyperparameter studies and templates can be found in Appendix~\ref{app:hyper} and~\ref{app:template}.
% Due to such reason, we conduct the following hyperparameter study and case studies on \textbf{Few-NERD} dataset.

\subsection{Case Study}\label{sec:case}
We further showcase some new instances generated based on few-shot examples in \textbf{Few-NERD} in Table~\ref{tab:new_inst}.
Four single-token instances and seven multiple-token instances are shown respectively. The generated results are of reasonable correctness and rich variety, thus ensuring the quality of the newly added training data.
We can also found that the generator is able to capture in-context instances like ``Zeus'', ``Hades'' in the first example and ``3DS'' in the last example, which are also important to improve the context-based entity typing model.

\section{related work}
This work focuses on few-shot entity typing using prompt-based methods, thus we will introduce two directions of work, prompt-based tuning and fine-grained entity typing.

\smallskip
\noindent \textbf{Prompt-Based Tuning.}~
PLMs~\cite{devlin2019bert,liu2019roberta, clark2020electra} have shown remarkable performance through fine-tuning on downstream tasks~\cite{Howard2018UniversalLM}, thanks to their learned generic linguistic features~\cite{Petroni2019LanguageMA} via pre-training. Some studies also show that PLMs are able to capture factual knowledge~\cite{Petroni2019LanguageMA} by making predictions on manually curated ``fill-in-the-blank'' cloze-style prompts.
Researchers use manually created prompts as task descriptions on GPT-3~\cite{Brown2020LanguageMA} and found that prompts can guide the model to generate answers for specific tasks, especially effective on few-shot tasks.
Inspired by these observations, studies on prompt engineering~\cite{Jiang2020HowCW, Shin2020ElicitingKF} aim to automatically search for optimal templates on specific tasks.
% and prompt-tuning~\cite{Lester2021ThePO, Li2021PrefixTuningOC} is further introduced to replace discrete tokens with continuous vectors for optimality.
In few-shot settings where training data is very limited, prompt-based tuning has surpassed standard model fine-tuning in a wide range of applications including text classification~\cite{Han2021PTRPT,Hu2021KnowledgeablePI}, relation extraction~\cite{Chen2021KnowPromptKP}, named entity recognition~\cite{Chen2021LightNERAL, Lee2021GoodEM}, and fine-grained entity typing~\cite{NNN2021PromptLearningFF}.

\smallskip
\noindent \textbf{Fine-Grained Entity Typing.}~
The goal of fine-grained entity typing (FET) is to determine the type of entity given a particular context and a label hierarchy~\cite{Ling2012FineGrainedER, Yosef2012HYENAHT}.
Several early studies~\cite{Ling2012FineGrainedER,Gillick2014ContextDependentFE,Dai2020ExploitingSR} generate large training data by automatically labeling entity mentions using external knowledge bases, which was followed by a line of research known as distantly-supervised FET with the goal of denoising the automatically generated labels~\cite{Ren2016LabelNR}. Some studies focus on leveraging type hierarchy and type inter-dependency~\cite{Ren2016AFETAF,Chen2020HierarchicalET,Murty2018HierarchicalLA,Lin2019AnAF} for noise reduction, while other studies~\cite{Jin2019FineGrainedET,Dai2019ImprovingFE,Xin2018ImprovingNF,Liang2020BONDBO,Meng2021DistantlySupervisedNE, Zhang2020LearningWN} combine external entity information provided in knowledge bases and self-training techniques to automatically relabel the data. Recently, joint training with relation extraction~\cite{Yaghoobzadeh2017NoiseMF} and searching for optimal templates under prompt-based framework~\cite{Dai2021UltraFineET} are also proposed in the distantly-supervised setting.

Zero-shot FET has also been explored to mitigate human annotation burden by transferring the knowledge learned from seen types to unseen ones. Multiple sources of information are used for zero-shot FET: \cite{Yuan2018OTyperAN} maps mention embedding to type embedding space by training a neural model to combine entity and context information.
\cite{Zhang2020MZETMA} captures hierarchical relationship between unseen types and seen types in order to generate unseen label representations.
\cite{Ma2016LabelEF} further enhances label representations by incorporating hierarchical and prototypical information derived from around 40 manually selected context-free entities as prototypes for each type.
\cite{Zhou2018ZeroShotOE, Obeidat2019DescriptionBasedZF} define unseen types by generating type embeddings from Wikipedia descriptions.
\cite{Chen2021AnES} fuses three distinct types of sources: contexutal, pre-defined hierarchy and label-knowledge (label prototype and descriptions), by training three independent modules to combine their results.
Nonetheless, these zero-shot learning algorithms require extensive annotations on source domain or manually selected high-quality representative mentions.

Our work focuses on the few-shot learning scenario, where each entity type is given a few annotated entity mention within a context, and no manual descriptions for type labels are provided. \cite{NNN2021PromptLearningFF} is a recent work that first explores prompt-based tuning on FET, and their setup is mostly similar to ours.

% demonstration-based NER \cite{Lee2021GoodEM}
% prompt-based  template + verbalizer
% verbalizer from knowledge graph\cite{Hu2021KnowledgeablePI}

\section{Conclusion and Future Work}
In this paper, we study the problem of few-shot fine-grained entity typing, where only a few contexutalized entity mentions are given for each type. We design two modules to empower prompt-based tuning approaches for this task. An entity label interpretation module is proposed to close the gap between the MLM head output space and the type label space by automatically learning word-type correlations from few-shot samples and the label hierarchy. To further utilize the generation power of PLMs, we propose a type-based contextualized generator to generate new instances based on the few-shot instances and their types, and use the new instances to generalize the initial training set. Our method outperforms previous baselines by large margins on three benchmark datasets.

There are several interesting directions for future studies: The type-based instance generator can benefit other entity-related tasks such as relation extraction; the label interpretation can improve other few-shot NLP tasks with prompt-based approaches. Moreover, one may also considers exploring generating new instances by ensembling from multiple large-scale language models.

\begin{acks}
Research was supported in part by Microsoft Research PhD Fellowship, Google PhD Fellowship, US DARPA KAIROS Program No. FA8750-19-2-1004 and INCAS Program No. HR001121C0165, National Science Foundation IIS-19-56151, IIS-17-41317, and IIS 17-04532, and the Molecule Maker Lab Institute: An AI Research Institutes program supported by NSF under Award No. 2019897, and the Institute for Geospatial Understanding through an Integrative Discovery Environment (I-GUIDE) by NSF under Award No. 2118329. Any opinions, findings, and conclusions or recommendations expressed herein are those of the authors and do not necessarily represent the views, either expressed or implied, of DARPA or the U.S. Government. The views and conclusions contained in this paper are those of the authors and should not be interpreted as representing any funding agencies.
\end{acks}

\bibliographystyle{ACM-Reference-Format}
\bibliography{acmart.bib}

\clearpage
\appendix
\onecolumn
\section{Hyperparameter Study.}~\label{app:hyper}
\begin{figure*}[h]
\subfigcapmargin=5pt
\centering
\subfigure[Accuracy on Few-NERD dataset with different smoothing parameter $\epsilon$ and initialization bias $\alpha$ applied.]{
	\label{fig:epsilon}
	\includegraphics[width = 0.35\textwidth]{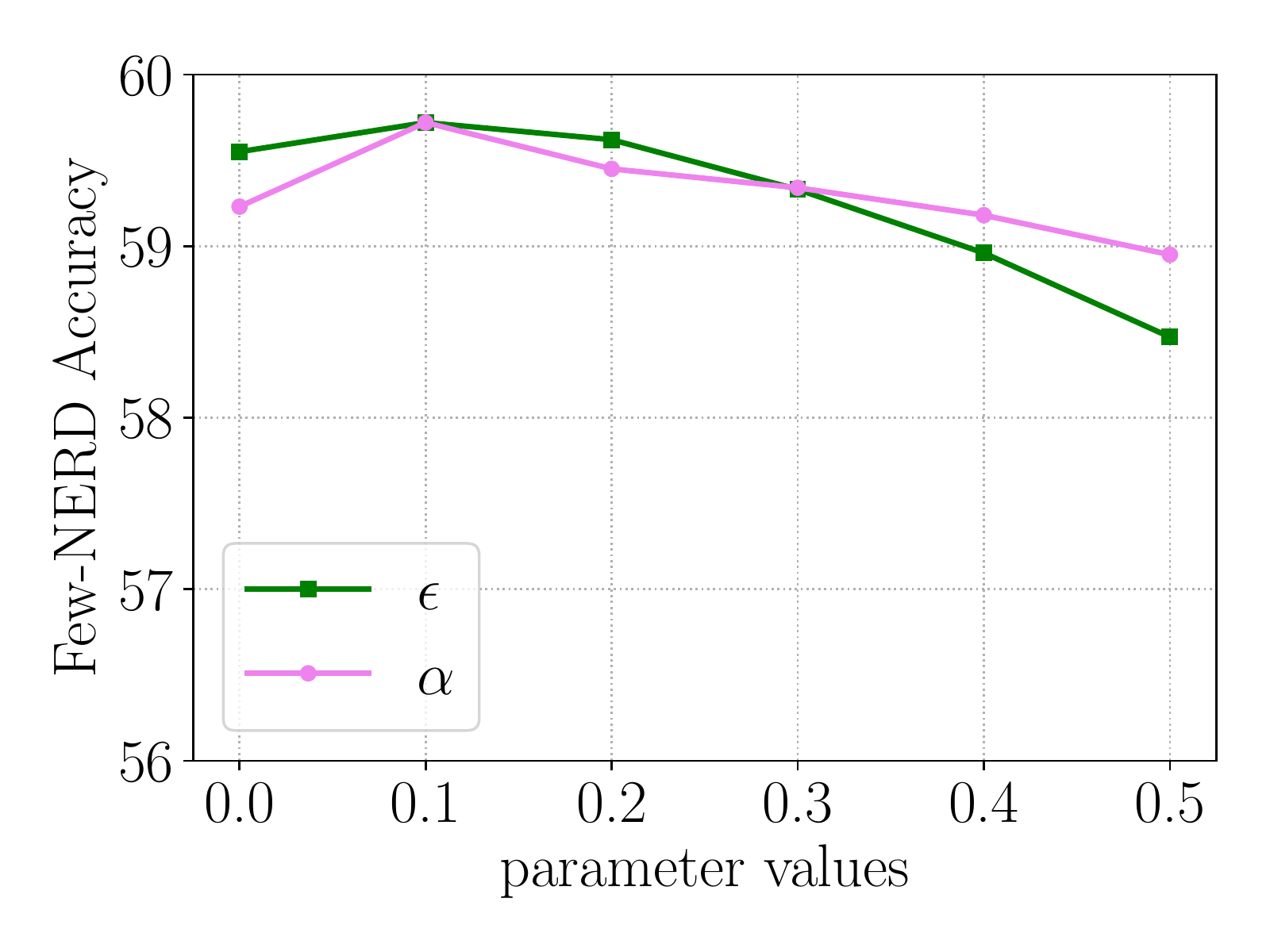}
}
\subfigure[Accuracy on Few-NERD dataset with different numbers of instance generated per type.]{
	\label{fig:num_inst}
	\includegraphics[width = 0.35\textwidth]{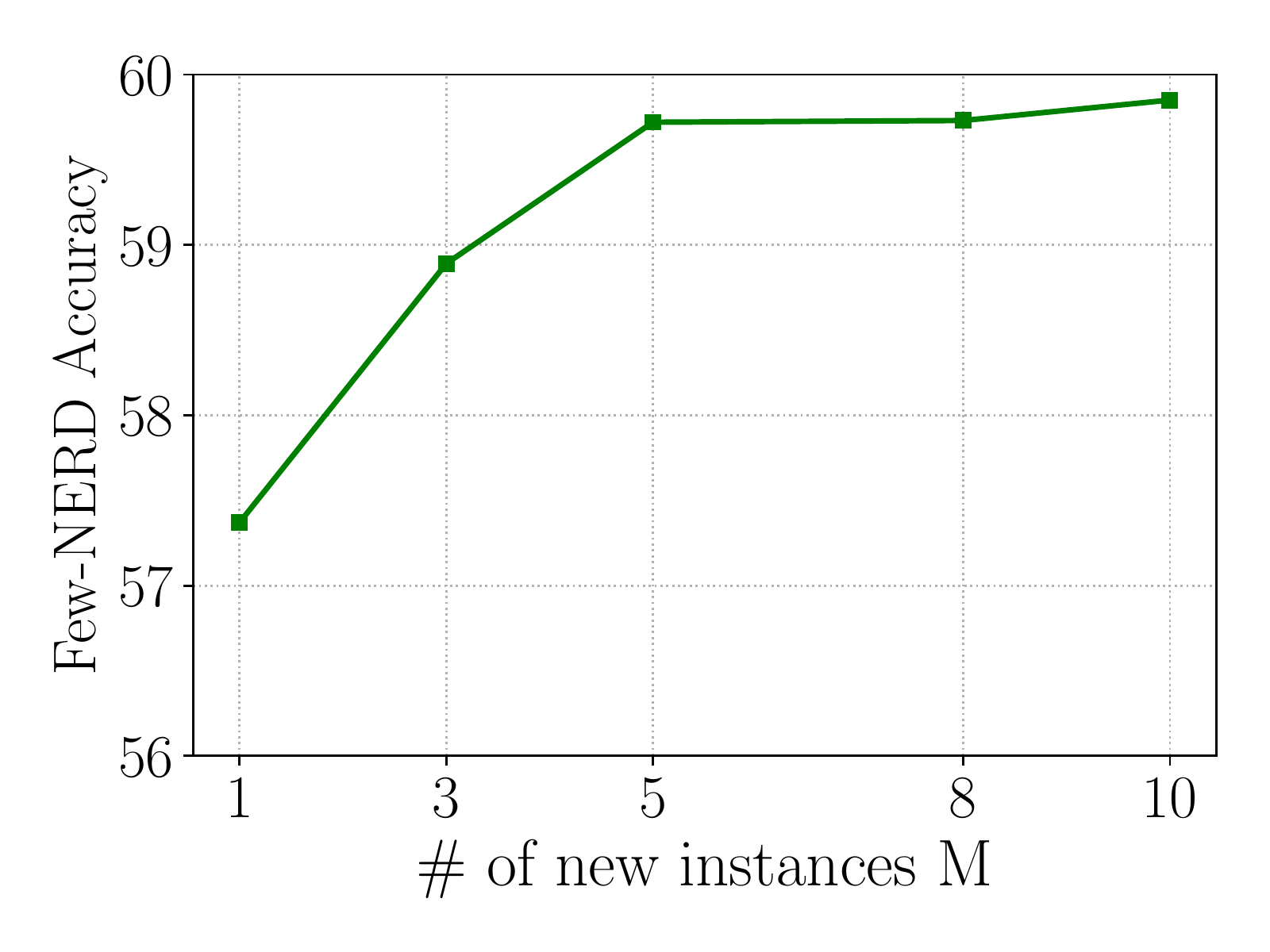}
}
% \subfigure[Testing Curve on Few-NERD dataset with or without type-based new instance generation module.]{
% 	\label{fig:curve}
% 	\includegraphics[width = 0.315\textwidth]{fig/w_instance.pdf}
% }
\caption{Hyperparameter study and testing curve analysis.
}
\label{fig:hyperparameter}
\end{figure*}

We study the effect of several hyperparameters in \model. We conduct all the hyperparameter study and case studies on \textbf{Few-NERD} dataset.
We first separately vary the initialization bias $\alpha$ in Eq.~\eqref{eq:init_U} and the label smoothing parameter $\epsilon$ in Eq.~\eqref{eq:label_smooth} in range $[0.0, 0.5]$, while keeping the other's value as default ($0.1$ for both default values). We report the accuracy on \textbf{Few-NERD} in Fig.~\ref{fig:epsilon}. Overall, the performance is insensitive to $\alpha$ since the values in correlation matrix $U$ are trainable. The accuracy also does not change much when $\epsilon$ is reasonably close to $0$, and can degrade a little when too much weight is given away from the original label.
We then vary the number of new instances $M$ generated added to training, and run experiments over $M=[1, 3, 5, 8, 10]$. We present the results in Fig.~\ref{fig:num_inst} and found that the model performance increases as more new instances are generated. However, the increasing trend slows down at $M=5$.
Since generating more new instances cost extra time, it is already sufficient for us to generate $5$ instances per type during training.

% \smallskip
% \noindent \textbf{Testing Curve.}~
% We compare the performance curve on the test set of \textbf{Few-NERD} with and without the type-based instance generator in Fig.~\ref{fig:curve}. With the proposed module generalizing the dataset at the second half of training (from epoch $11$ to $20$), the model continues to learn from the enhanced dataset and converges much slower and at a higher point than without the module.

\section{Effects of Templates.}~\label{app:template}
Previous studies~\cite{Gao2021MakingPL, Zhao2021CalibrateBU} point out that the quality of prompt templates play an important role in the performance of prompt-based methods. To study the impact of different template choices, we replace our template with two more templates and show their performance in Table~\ref{tab:template}. The results indicate that the choice of templates can impact the final performance, and the result is better when the typing template $T_c$ matches more with the generating template $T_g$.

\begin{table*}[h]
\centering
\caption{Performance with different templates on the Few-NERD dataset.}\label{tab:template}
\begin{tabular}{c|c|c|c|c}
\toprule
Typing template $T_c(\xv,\mv)$ & Instance generating template $T_g(\xv, \mv, t)$ & Acc. & Micro-F1 & Macro-F1 \\
\midrule
$\xv$. $\mv$ is a [MASK]. & $\xv$. $\mv$, as well as [MASK], is a $t$. & 59.72 & 71.90 & 71.90\\
$\xv$. In this sentence, $\mv$ is a [MASK]. & $\xv$. $\mv$, as well as [MASK], is a $t$. & 58.57 & 71.69 & 71.69\\
$\xv$. $\mv$ is a type of [MASK]. & $\xv$. $\mv$, as well as [MASK], is a type of $t$. & 58.86 & 72.05 & 72.05\\
\bottomrule
\end{tabular}
\end{table*}

\end{document}